\documentclass[conference,9pt]{IEEEtran}
\IEEEoverridecommandlockouts
\usepackage{cite}
\usepackage{amsmath,amssymb,amsfonts}
\usepackage{algorithmic}
\usepackage{graphicx}
\usepackage{textcomp}
\usepackage{xcolor}
\usepackage{algorithm}
\usepackage{booktabs}
\usepackage{multirow}
\usepackage{array}
\usepackage{colortbl}
\usepackage{makecell}
\usepackage{pifont}
\usepackage{makecell}
\usepackage{xcolor,colortbl}

\def\BibTeX{{\rm B\kern-.05em{\sc i\kern-.025em b}\kern-.08em
    T\kern-.1667em\lower.7ex\hbox{E}\kern-.125emX}}

    \colorlet{mygreen}{green!60!gray}

    \newif\ifcomments

\commentstrue

\ifcomments

    \newcommand{\BTcomm}[1]{\textcolor{mygreen}{{#1}}}
    \newcommand{\MB}[1]{\textcolor{magenta}{{MB: #1}}}
    \newcommand{\fjwresp}[1]{\textcolor{blue}{#1}}
    \newcommand{\fjwrespv}[1]{\textcolor{red}{#1}}

\else
    \newcommand{\BTcomm}[1]{}
    \newcommand{\MB}[1]{}
    \newcommand{\fjwresp}[1]{}
    \newcommand{\fjwrespv}[1]{}

\fi

\begin{document}

\title{Robust Retraining-free GAN Fingerprinting via Personalized Normalization}

\author{
\IEEEauthorblockN{Jianwei Fei}
\IEEEauthorblockA{\textit{Jinan University} \\
Guangzhou, China}
\and
\IEEEauthorblockN{Zhihua Xia \thanks{ Zhihua Xia is the corresponding author.}\IEEEauthorrefmark{2}}
\IEEEauthorblockA{\textit{Jinan University} \\
Guangzhou, China}
\and
\IEEEauthorblockN{Benedetta Tondi}
\IEEEauthorblockA{\textit{University of Siena} \\
Siena, Italy}
\and
\IEEEauthorblockN{Mauro Barni}
\IEEEauthorblockA{\textit{University of Siena} \\
Siena, Italy}
}

\maketitle

\begin{abstract}
In recent years, there has been significant growth in the commercial applications of generative models, licensed and distributed by model developers to users, who in turn use them to offer services. In this scenario, there is a need to track and identify the responsible user in the presence of a violation of the license agreement or any kind of malicious usage. Although there are methods enabling Generative Adversarial Networks (GANs) to include invisible watermarks in the images they produce, generating a model with a different watermark, referred to as a fingerprint, for each user is time- and resource-consuming due to the need to retrain the model to include the desired fingerprint. In this paper, we propose a retraining-free GAN fingerprinting method that allows model developers to easily generate model copies with the same functionality but different fingerprints. The generator is modified by inserting additional Personalized Normalization (PN) layers whose parameters (scaling and bias) are generated by two dedicated shallow networks (ParamGen Nets) taking the fingerprint as input. A watermark decoder is trained simultaneously to extract the fingerprint from the generated images. The proposed method can embed different fingerprints inside the GAN by just changing the input of the ParamGen Nets and performing a feedforward pass, without finetuning or retraining. The performance of the proposed method in terms of robustness against both model-level and image-level attacks is also superior to the state-of-the-art. 
\end{abstract}


\section{Introduction}
Synthetic image generation has made significant progress in recent years and generative models are now widely used in commercial applications. These models are provided to commercial users as production tools or for selling services. Protecting the Intellectual Property Rights (IPR) of model owners has become a pressing issue to avoid potential copyright infringements, such as unauthorized duplication or model theft, when these models are delivered to malicious users. Deep Neural Network (DNN) watermarking has been proposed as a solution to protect the IPR associated with DNN models~\cite{barni2021dnn}. Most DNN watermarking methods focus on the protection of discriminative models, namely, networks developed for classification tasks, and less attention is paid to generative models. Yet, some methods for the watermarking of generative models have started appearing recently. Given the large entropy of the output of generative models, the watermark can be directly extracted from the output produced by the model, thus permitting to verification of the watermark in a so-called {\em box-free} setting. In this way, it is possible to determine the source of images produced by generative models and associate any image to the generative model that produced it~\cite{ong2021protecting}.

\begin{figure}[thbp]
    \centering
    \includegraphics[width=\linewidth]{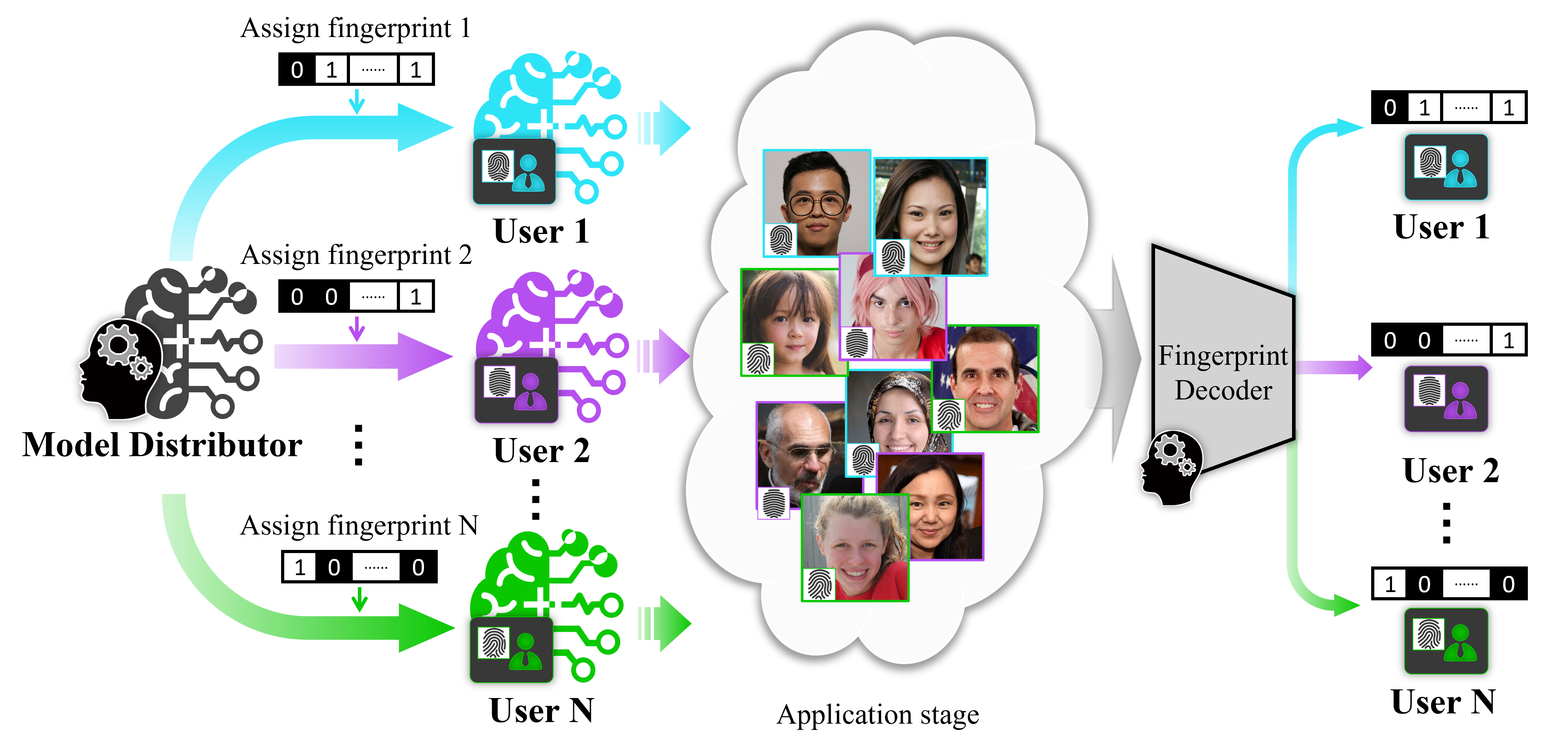}
    \caption{The GAN fingerprinting scenario considered in this paper.}
     \label{fig:fig1}
     \vspace{-1.0em}
\end{figure}

With the exception of a few scattered works~\cite{DBLP:conf/iclr/YuSCDF22}, the existing approaches for the watermarking of generative models, notably GANs, are designed for ownership verification, aiming at making it possible to retrieve the model authorship information from the generated images. These methods embed a fixed watermark, linking the model to the owner, and require retraining or finetuning if different a watermark has to be embedded in the model. 

In this paper, we focus on a different scenario, hereafter referred to as GAN fingerprinting, illustrated in Fig.~\ref{fig:fig1}. In this scenario, the model distributor, simply referred to as the model owner, releases distinct watermarked model instances to different users, in such a way that the user-specific fingerprint can be recovered from the images produced by these models for copyright authentication and to trace back to the guilty user in case of a violation of the license agreements (traitor tracing). A problem with this scenario is that the model owner must produce several instances of the model each containing a different watermark, in this case referred to as a fingerprint, which in general requires training or finetuning a new model for each user. In this paper, we propose a retraining-free GAN fingerprinting method for box-free watermarking of GAN models, permitting to {\em easily} create different model instances each containing a different fingerprint, without any need to retrain or finetuning the watermarked model. This goal is achieved by introducing a new Personalized Normalization (PN) layer into the generator's architecture, whose scaling factor and bias parameters are determined by two independent shallow networks, called the Parameters Generation Networks (ParamGen Nets), that are fed with the input fingerprint sequence. The generator and the ParamGen Nets are trained jointly with the watermark decoder, by varying the fingerprint sequence during training across the iterations. Once the model is trained, the model distributor can easily get GAN models with different fingerprints by just changing the fingerprint sequence at the input of the ParamGen Nets and performing a feedforward pass to modify the parameters of the PN layers of the GAN accordingly. Our idea of embedding the watermark information in the normalization layers is inspired by passport-based approaches~\cite{zhang2020passport} proposed for the protection of the IPR of DNN classifiers, which use passports, namely digital signatures, to unlock the normalization layer and use the network at inference time, achieving remarkable robustness against both removal and ambiguity attacks. Robustness is a crucial requirement for DNN watermarking algorithms aimed at IPR protection~\cite{barni2021dnn}. Malicious users can conduct watermark removal attacks by model-level modifications, moreover, box-free GAN watermarking is also subject to image-level attacks. An additional strength of our method is that the watermark embedded in the GAN models is a very robust one. In particular, the experiments we run show that good robustness is achieved against both model-level attacks, specifically finetuning and model compression (pruning and quantization), and image post-processing operations, like JPEG compression, noise addition, and Gaussian blur.

To the best of our knowledge, the only work proposing a retraining-free, GAN fingerprinting method is~\cite{DBLP:conf/iclr/YuSCDF22}, which uses the fingerprint sequence to modulate the parameters of the convolutional kernels. Compared to~\cite{DBLP:conf/iclr/YuSCDF22}, our approach can achieve improved robustness against both image-level and model-level removal attacks. We believe that this gain comes from the different approach we are considering for watermark embedding, namely the PN-based embedding.

The rest of the paper is organized as follows. In Section~\ref{Related Work}, we briefly discuss the state of the art of GAN watermarking. The proposed method is described in Section~\ref{Efficient GAN Fingerprinting}. Section~\ref{Experimental Evaluation} reports the experimental methodology, settings, and results. Finally, in Section~\ref{Conclusions}, we conclude the paper with some final remarks and hints for future research.

\section{Related Work}
\label{Related Work}
The goal of DNN watermarking is to embed watermarks into deep learning networks, that can be used for ownership verification, fingerprinting, and traitor tracing, among other possible applications, without impairing their functionality (unobtrusiveness)~\cite{li2021survey,barni2021dnn}. While DNN watermarking has been mostly applied to CNN based classifiers, the watermarking of GAN has recently started receiving attention~\cite{wu2020watermarking, ong2021protecting, yu2021artificial,fei2022supervised}. Depending on the kind of access required for the watermark extraction, watermarking algorithms can be categorized as white-box, black-box, and box-free. White-box methods require access to the internal parameters of the model during verification. With black-box methods, instead, the watermark is extracted by looking at the output of the network in correspondence to a set of specific querying inputs. Finally, box-free watermarking methods do not require any kind of access to the suspicious model, and the watermark can be directly extracted from the output produced by the model. Box-free methods are only possible with generative models, for which case the entropy of the output is large enough. Black-box and box-free methods have been proposed for GAN watermarking. In~\cite{ong2021protecting,qiao2023novel,quan2020watermarking}, the authors utilize black-box watermarking to protect the IPR of generative models where the watermark is embedded by instructing the model to learn to produce specific output images when fed with certain inputs. Several methods have been provided performing box-free watermarking~\cite{yu2021artificial, fei2022supervised, wu2020watermarking}. In particular, Wu~\textit{et al.}~\cite{wu2020watermarking} imposed an additional constraint on the output of the generator in the loss, in such a way as to embed a specific watermark image into any generated image. This method is a zero-bit watermarking that utilizes the PSNR between the extracted image and the ground-truth for ownership verification. Yu~\textit{et al.}~\cite{yu2021artificial} discover that by training the GAN using data with a watermark embedded via StegaStamp~\cite{tancik2020stegastamp}, the generated images will also contain the watermark. Fei~\textit{et al.}~\cite{fei2022supervised} propose to improve the method in~\cite{yu2021artificial} by embedding the watermark in a supervised manner, using a pre-trained watermark decoder to guide the training and incorporating a loss term in the optimization to make sure that the watermark extracted from the generated images is close to the true watermark.

The above watermarking methods are developed with the ownership verification application in mind. In fingerprinting applications, a company may want to distribute to different users model instances having the same functionality but with distinct (user-specific) fingerprints. In this scenario, the above box-free watermarking methods would require retraining a new generative model for every user, with enormous costs. To address this problem, Yu~\textit{et al.}~\cite{DBLP:conf/iclr/YuSCDF22} propose a method for the efficient fingerprinting of GAN models. This method enables the efficient generation of model instances with the same image generation functionality with different user-specific fingerprints, achieved by exploiting watermark autoencoders and modulating the parameters of the convolutional kernels based on the to-be-embedded watermark. To the best of our knowledge, \cite{DBLP:conf/iclr/YuSCDF22} is the only one proposing a method for retraining-free GAN fingerprinting. As a drawback, this method has limited robustness against network modification and re-use (model-level attack) and against image post-processing (image-level attacks), which hinders its practical applicability.

In this paper, we propose a retraining-free GAN fingerprinting method with improved robustness against both model-level and image-level attacks. In particular, the robustness against finetuning is largely improved with respect to~\cite{DBLP:conf/iclr/YuSCDF22}. The cost in terms of time and resources is the same as in~\cite{DBLP:conf/iclr/YuSCDF22}, since both methods do not need any retraining and a user-specific fingerprint can be embedded by feeding the ParamGen Nets with the fingerprint, namely the watermark message, and using the resulting parameters in the PN layer running a feedforward pass thought the generation network to set the parameters of the generator instance associated to the user. Table~\ref{table:comparisons} summarizes the capabilities of state-of-the-art GAN watermarking approaches in terms of capacity, efficiency, and robustness. The term efficiency here refers to the capability of changing the fingerprint, without retraining or finetuning the model. Our method is the only one with all the desired capabilities.

\vspace{-1.0em}
\begin{table}[htbp]
    \renewcommand{\arraystretch}{1}
    \centering
    \caption{Comparisons of state-of-the-art box-free GAN watermarking methods and the proposed method.}
    \begin{tabular}{cccc}
    \toprule
    Approach                         & Capacity   &  Retraining-free    & Robustness  \\ \midrule
    Wu \textit{et al.} \cite{wu2020watermarking}  & Zero-bit   &  \ding{56} & \ding{52}   \\
    Yu \textit{et al.} \cite{yu2021artificial}         & Multi-bit  &  \ding{56} & \ding{56}   \\
    Fei \textit{et al.} \cite{fei2022supervised}      & Multi-bit  &  \ding{56} & \ding{56}  \\
    Yu \textit{et al.} \cite{DBLP:conf/iclr/YuSCDF22}   & Multi-bit  &  \ding{52} & \ding{56}  \\
    Ours                           & Multi-bit  & \ding{52}  & \ding{52}  \\
    \bottomrule
    \end{tabular}
    \label{table:comparisons}
\end{table}

\section{Proposed Method}
\label{Efficient GAN Fingerprinting}

In this section, we describe the proposed approach for robust retraining-free GAN fingerprinting. In its simplest form, a GAN model consists of a generator and a discriminator, denoted by $G$ and $D$, respectively. $G$ takes a noise sample $z \in \mathbb{R}^{d_z} \sim P_z$ of dimension $d_z$ as input and produces a synthetic image at the output. $G$ and $D$ are updated alternatively during training. The loss that $D$ wants to maximize can be expressed as:
\begin{equation}
    \mathcal{L}_{D}=\underset{x \sim p_{\text {x}}}{\mathbb{E}} \log D(x)+ \underset{\substack{z \sim P_z \\ w \sim \{0,1\}^{d_w}}}{\mathbb{E}} \log (1-D(G(z))),
    \label{eq:adv loss-D}
\end{equation}
where $x$ denotes the real image and $p_{\text {x}}$ its distribution, while $G$ wants to minimize
\begin{equation}
    \mathcal{L}_{G}=\underset{\substack{z \sim P_z \\ w \sim \{0,1\}^{d_w}}}{\mathbb{E}} \log (1-D(G(z))).
    \label{eq:adv loss-G}
\end{equation}

\begin{figure}[thbp]
    \centering
    \includegraphics[width=0.9\linewidth]{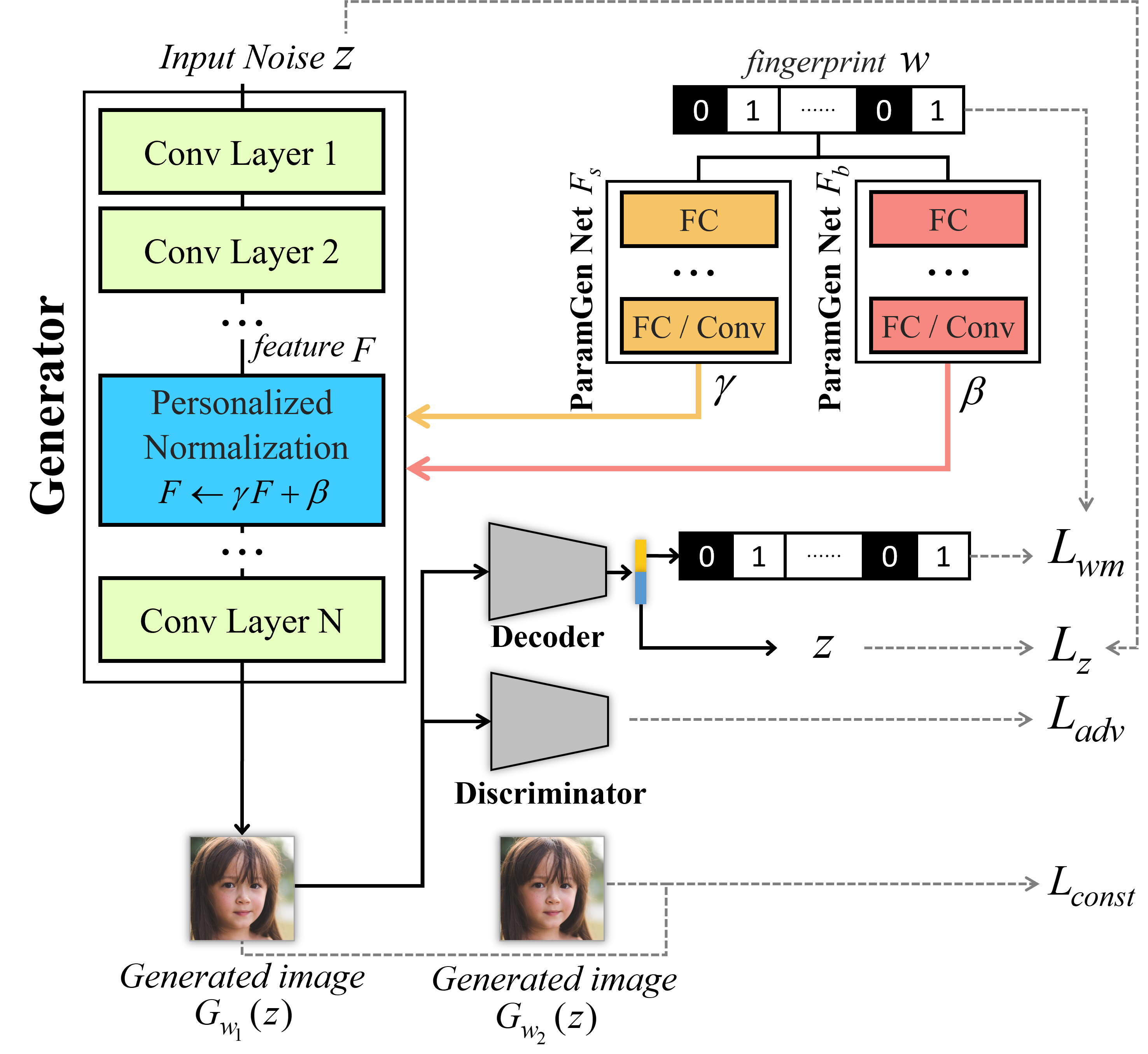}
    \caption{Overview of the proposed GAN fingerprinting approach. Green blocks in the generator represent the original layers, while the blue block represents the PN layer added for watermark embedding. 
    }
     \label{fig:overview}
     \vspace{-0.3cm}
\end{figure}

To watermark the generator $G$, we insert an additional intermediate layer performing personalized normalization and consider two (trainable) parameter generation networks $F_s$ and $F_b$, named ParamGen Nets, taking as input the watermark message $w$ with $d_w$ bits ($w \in \{0,1\}^{d_w}$) and producing respectively the scaling factor $\gamma$ and bias $\beta$ used in the PN layer of the generator\footnote{The ParamGen Nets are described in details in Sect.~\ref{sec.paramNorm}. Depending on the variant of the method (two variants are considered, see Sect.~\ref{sec.paramNorm}), $\gamma$ and $\beta$ can be either vectors or tensors.}, and a (trainable) watermark decoder $D_{w}$. We denote with $G_w$ the generator $G$ with the PN layer parameterized by $w$. The overall architecture is shown in Fig.~\ref{fig:overview}. A model with the desired retraining-free fingerprinting functionality is achieved by training the architecture as described below, by introducing three new loss terms for training.

To instruct the network to embed a different watermark inside the images it produces for every different $w$, the watermark $w \in \{0,1\}^{d_w}$ is randomly sampled and $D_{w}$ is jointly trained to extract the watermark from $G_w(z)$, in such a way to minimize the bit-wise error between the watermark $w$ and the output of the decoder, that is
\begin{equation}
    \begin{aligned}
     \mathcal{L}_{wm} = & \underset{\substack{z \sim P_z \\ w \sim \{0,1\}^{d_w}}}{\mathbb{E}} \sum_{i=1}^{d_w} w_i \log \sigma ((D_w(G_w(z)))_i)  \\ & + (1 -w_i) \log (1-\sigma ((D_w(G_w(z)))_i)),
    \end{aligned}
    \label{eq:watermark loss}
\end{equation}
where $\sigma$ is the sigmoid function and $(D_w(G_w(z)))_i$ denote the $i$-th element of $D_w(G_w(z))$.

To ensure that the content of the generated image $G_w(z)$ is regulated by the input noise $z$, by following~\cite{DBLP:conf/iclr/YuSCDF22} and~\cite{srivastava2017veegan}, we instruct the decoder $D_w$ to recover the input noise $z$ in addition to extracting the watermark $w$. Therefore, the output of $D_w$ is a vector of length ${d_w + d_z}$ where the first $d_w$ elements are dedicated to the watermark (Eq.~\ref{eq:watermark loss}) and the last $d_z$ elements are used for input recovery. The above goal is achieved by minimizing the $L_2$ reconstruction loss
\begin{equation}
    \mathcal{L}_z=\underset{\substack{z \sim P_z \\ w \sim \{0,1\}^{d_w}}}{\mathbb{E}} \sum_{i=1}^{d_z}\left(z_i- (D_{w}(G_w(z)))_{d_w+i}\right)^2.
    \label{eq:z reconstruction loss}
\end{equation}
Finally, to ensure that various model instances behave in the same way, regardless of the embedded watermark, we impose an additional constraint requiring that for the same sample noise $z$ the generators parameterized by different $w$ results in the same output image. Following~\cite{DBLP:conf/iclr/YuSCDF22}, this goal is achieved via an image consistency loss
\begin{equation}
    \mathcal{L}_{\text {const }}=\underset{\substack{z \sim P_z \\ w_1, w_2 \sim \{0,1\}^{d_w}}}{\mathbb{E}} \left\|G_{w_1}\left(z\right)-G_{w_2}\left(z\right)\right\|_2^2,
    \label{eq:image consistency loss}
\end{equation}
where $w_1$ and $w_2$ are two distinct watermarks sampled randomly during training. Eventually, the overall loss used to train the watermarked generator is:
\begin{equation}
    \mathcal{L}_{G,tot} = \lambda_1 \mathcal{L}_{G_w} + \lambda_2 \mathcal{L}_{wm} + \lambda_3 \mathcal{L}_z + \lambda_4 \mathcal{L}_{\text {const}},
    \label{eq:total loss}
\end{equation}
where $\lambda_1$, $\lambda_2$, $\lambda_3$, and $\lambda_4$ weight each term of the loss (and $\mathcal{L}_{G_w}$ is the loss defined in \eqref{eq:adv loss-G} with $G_w$ in place of $G$). The loss of $D$ is the same as in \eqref{eq:adv loss-D} with $G_w$ replacing $G$.

Once the GAN fingerprinting model is trained, given a new user $m$, the model $G_{w_m}$ containing the user-specific fingerprint $w_m$ is obtained, by assigning PN layers the parameters obtained through ParamGen Nets. The generator is then distributed to the user, while $F_s$, $F_b$, and $D_{w}$ are kept by the model owner.

\subsection{Personalized Normalization}
\label{sec.paramNorm}
In this section, we provide the details of the ParamGen Nets $F_s$ and $F_b$. These are two independent networks that take as input $w$ and output $\gamma = F_s(w)$ and $\beta  = F_b(w)$, that are used to normalize the feature map $F$ at the input of the PN layer. Normalization is performed in two different ways, referred to as channel-wise PN and element-wise PN, with the output of the ParamGen Nets $\gamma$ and $\beta$ having different dimensionality in the two cases.

\subsubsection{Channel-wise PN}
The output of the ParamGen Nets are $\gamma \in \mathbb{R}^c$,  $\beta \in \mathbb{R}^c$, where $c$ is the number of channels of the feature map $F$, and $F_s$ and $F_b$ are fully-connected (FC) networks.\footnote{The exact architecture of the ParamGen Nets is provided in the methodology section.} Then, $F$ is scaled and translated by channels and we have
\begin{equation}
     F'_{ijk} = \gamma_k  F_{ijk} + \beta_k,
    \label{eq:cw2}
\end{equation}
for $i \in[1, p], j \in[1, q], k \in[1, c]$, where $p$ and $q$ are the height and width of $F$.

\subsubsection{Element-wise PN}
In this case, $\gamma \in \mathbb{R}^{p \times q \times c}$ and $\beta \in \mathbb{R}^{p \times q \times c}$ produced by two convolutional networks $F_s$ and $F_b$ that consists of an input FC layer followed by convolutional layers. Then, $F$ is scaled and translated by the corresponding entries of $\gamma$ and $\beta$, that is, $F' = \gamma \circ F + \beta$, where $\circ$ denotes the element-wise product.

\section{Experimental Methodology and Analysis}
\label{Experimental Evaluation}

\subsection{Methodology and Settings}
\label{Experimental Setting}

\textbf{Models and dataset.} We run experiments on several GAN architectures, focusing on the generation of face images. Specifically, we consider the following networks: Boundary Equilibrium GAN (BEGAN)~\cite{berthelot2017began2}, Spectral Normalization GAN (SNGAN)~\cite{miyatospectral}, and Progressive Growing GAN (PGGAN)~\cite{karrasprogressive}. As for the pristine set, we consider 200k images with 64$\times$64 resolution from CelebA~\cite{liu2015deep}. The models are trained using the official code, but the architecture and training procedure are modified to implement the proposed GAN fingerprinting approach.

\textbf{Architecture and training.}
In the proposed GAN fingerprinting architecture, the PN layer is added as the penultimate layer of the generator (default). Other positions for the PN layer are considered in the ablation study presented in Sect.~\ref{Ablation study}. The details of the ParamGen Nets are provided in Table~\ref{table: Architectures information} for both channel-wise and element-wise cases. For every layer (row), the type of layer, input dimension, output dimension, and the type of activation are reported. Regarding the parameters, we set $p$ and $q$ for the element-wise PN equal to $32$, while $c$, which corresponds to the number of channels in the feature maps, takes a different value for the various architectures. In particular, $c = 128$, $64$, and $128$, respectively for BEGAN, SNGAN, and PGGAN. During training, the watermarks are randomly selected for each sample in every batch, hence every image has a different watermark message associated to it. The weights are set as follows: $\lambda_i = 1$ for $i = 1,2,4$, $\lambda_3 = 0.1$. All models are trained for 50 epochs. In the case of PGGAN, which implements a progressive growing training procedure, blocks of layers are incrementally added during training, and the output size of the generator model (image resolution) progressively increases. Then, for this case, the generator is trained normally at the beginning (with the $\mathcal{L}_G$ and $\mathcal{L}_D$ loss). The PN layer and the watermark decoder are inserted only at a later stage, when the image resolution reaches 64$\times$64, and we started training the generator with the new loss\footnote{In order to work properly, the watermark decoder requires that the input size is fixed.}. In all the experiments, we set $d_w$ to 128, thus embedding a 128-bit fingerprint, that permits us to get $2^{128} \approx 3.4 \times 10^{38}$ distinct generator instances.

To increase the robustness of the watermark against image processing attacks, following~\cite{fei2022supervised}, we introduce a preprocessing layer before the watermark decoder. The considered processing includes JPEG compression with quality factors selected uniformly in [20, 50], Gaussian blur, with kernel size in [0, 9], and Gaussian noise addition with standard deviation in [0.001, 0.15]. Each processing is applied with probability equal to 15\%.

\begin{table}[htbp]
    \centering
    \caption{Structure of ParamGen Nets in the case of channel-wise PN (left) and element-wise PN (right).}
    \label{table: Architectures information}
    \begin{tabular}{c}
        \toprule
        FC, $d_w$, 512, ReLu        \\\midrule
        FC, 512, 512, ReLu          \\\midrule
        FC, 512, 512, ReLu              \\\midrule
        FC, 512, $c$, ReLu                  \\
        \bottomrule
    \end{tabular}
    \quad
    \begin{tabular}{c}
        \toprule
        FC, $d_w$, $8\times8\times32$, ReLu    \\\midrule
        Conv, $8\times8\times32$, $16\times16\times32$, ReLu    \\\midrule
        Conv, $16\times16\times32$, $32\times32\times64$, ReLu                  \\\midrule
        Conv, $32\times32\times64$, $p\times q \times c$, ReLu                        \\
        \bottomrule
    \end{tabular}
  \end{table}

\textbf{Metrics.} To measure the quality of the generated images, we use the Fr\'{e}chet Inception Distance (FID)~\cite{heusel2017gans}, commonly adopted in the literature. The FID scores of the generated images are calculated on a population of $5 \times 10^4$ real and $5 \times 10^4$ generated images, obtained from random generator instances, that is generators with random watermarks. To measure the effectiveness of watermark embedding, we use bit-wise accuracy (Acc) of watermark extraction, namely the percentage of bits that are correctly recovered. The bit-wise accuracy reported in the experiments is averaged on $10^4$ samples with random watermarks.

\subsection{Performance Analysis}
\label{Effectiveness}
The performance of our method for the various architectures is reported in Table~\ref{table:acc}, compared with the state-of-the-art box-free GAN watermarking approaches. In the table, -\textit{cw} and -\textit{ew} refer to channel-wise and element-wise PN layer. The 'No wm' column shows the baseline FID of the images generated by the non-watermarked model.

Regarding the quality, we observe that all methods get an FID similar to the baseline, thus the quality of the generated images is similar to that of the images generated by the non-watermarked GAN. The watermark accuracy is also nearly perfect in all the cases for all the methods, except for Yu~\textit{et al.}~\cite{yu2021artificial} (the lower effectiveness of this method is probably due to the fact that it performs embedding in an unsupervised manner, by simply training on watermarked images). The row 'Overhead' reports the time required by algorithms for the embedding of a new watermark. For the method in~\cite{yu2021artificial}, which requires retraining on the watermarked dataset, the time consumption is high. This value is high also for the method in~\cite{fei2022supervised}, which however only requires finetuning, with a lower overhead with respect to~\cite{yu2021artificial}. Yu~\textit{et al.}~\cite{DBLP:conf/iclr/YuSCDF22} and our method only requires hundreds of milliseconds resulting in a huge gain (in the order of $10^{4}$ to $10^{5}$). In our case, this is the time necessary for a single feedforward pass of the ParamGen Nets.
\begin{table}[htbp]
    \renewcommand{\arraystretch}{1.10}
    \centering
    \footnotesize
    \caption{Watermark accuracy, FID and Overhead for the various architectures.}
\begin{tabular}{m{0.9cm}<{\centering}|m{0.9cm}<{\centering}m{0.5cm}<{\centering}m{0.6cm}<{\centering}m{0.6cm}<{\centering}m{0.6cm}<{\centering}m{0.6cm}<{\centering}m{0.6cm}<{\centering}}
    \toprule
    Model    & Metric  & No wm & Yu~\cite{yu2021artificial}  & Fei~\cite{fei2022supervised}   & Yu~\cite{DBLP:conf/iclr/YuSCDF22}  &-\textit{cw} & -\textit{ew}\\ \midrule

    \multirow{3}*{BEGAN}    & Acc &  - & 93.69  & 99.10    & \textbf{100.00}  &99.87 & \textbf{100.00}\\
    & FID  & 20.89 & 25.12  & 20.84  & 21.78  &21.19  & \textbf{20.72}\\
    & Overhead   & - & 12h & 4h & \textbf{100}ms  &\textbf{100}ms & \textbf{100}ms\\ \midrule

    \multirow{3}*{SNGAN}    & Acc  & - & 92.77  & 99.45   & 99.99  & \textbf{100.00} & 99.99\\
    & FID   & 24.25 & 27.74 & 25.26  & 24.69 & \textbf{24.12} & 24.70\\
    & Overhead  & - & 8h  & 2h   & \textbf{100}ms  &\textbf{100}ms & \textbf{100}ms\\ \midrule

    \multirow{3}*{PGGAN}    & Acc  & - & 90.26  & 98.74   & 99.50 & \textbf{99.89} & 99.85 \\
    & FID   & 27.50 & 32.36  & 28.53    & 28.42 & 28.77 & \textbf{28.02} \\
    & Overhead   & -& 24h  & 8h   & \textbf{100}ms  &\textbf{100}ms & \textbf{100}ms\\

    \bottomrule
\end{tabular}
\label{table:acc}
\end{table}

\subsection{Robustness Analysis}
\label{sec.robustness}
We also evaluate the robustness of our GAN fingerprinting method against both model-level and image-level attacks.
\begin{table}[htbp]
    \renewcommand{\arraystretch}{1.10}
    \centering
    \caption{Performance (Acc/ FID) under different model-level attacks.
    }
\begin{tabular}{m{0.8cm}<{\centering}|m{1.2cm}<{\centering}m{1.1cm}<{\centering}m{1.1cm}<{\centering}m{1.1cm}<{\centering}m{1.1cm}<{\centering}}
    \toprule
    Model          & Approach          & \makecell[c]{Finetune\\(20k)}    & \makecell[c]{Prune\\(10\%)} & \makecell[c]{Prune\\(20\%)} &  \makecell[c]{Quant\\($10^{-1}$)}\\ \midrule

    \multirow{4}*{BEGAN} &Yu~\cite{DBLP:conf/iclr/YuSCDF22}       & 53.6/\textbf{20.6}  & 99.1/21.2 & 62.1/76.6 &   97.1/34.8\\
    &Fei~\cite{fei2022supervised}   & 56.5/20.7  & 99.1/\textbf{20.5}  & 61.7/82.2 &   98.0/\textbf{32.0}\\
    &Ours -\textit{cw}  & 75.9/20.9 &  99.5/22.0 & 66.6/\textbf{74.2} &  \textbf{98.8}/37.1 \\
    &Ours -\textit{ew}  & \textbf{85.0}/\textbf{20.6}  & \textbf{99.9}/21.9 & \textbf{68.8}/90.1&   98.7/36.7 \\ \midrule

    \multirow{4}*{SNGAN} &Yu~\cite{DBLP:conf/iclr/YuSCDF22}       & 60.0/24.3  & \textbf{99.9}/\textbf{24.8} & \textbf{82.6}/43.1 &   98.4/\textbf{26.2}\\
    &Fei~\cite{fei2022supervised}   & 62.0/24.1  & 98.0/26.3 & 80.2/45.7 &   96.0/26.7\\
    &Ours -\textit{cw}  & 84.5/\textbf{23.9} & 98.9/25.8 & 79.6/39.5 &  99.9/27.8 \\
    &Ours -\textit{ew}  & \textbf{88.2}/24.0  & 98.8/26.9 & 80.3/\textbf{39.1} &  \textbf{99.8}/28.6 \\ \midrule

    \multirow{4}*{PGGAN} &Yu~\cite{DBLP:conf/iclr/YuSCDF22}       & 64.2/28.4   & 98.9/29.6 & \textbf{88.4}/\textbf{37.1} &   99.1/30.1\\
    &Fei~\cite{fei2022supervised}   & 65.3/28.1  & 97.4/30.2 & 84.2/39.8 &   99.0/32.4\\
    &Ours -\textit{cw}  & 73.0/\textbf{27.2} &  \textbf{99.5}/29.6& 84.2/42.3 &  99.5/32.1 \\
    &Ours -\textit{ew}  & \textbf{74.5}/28.0   & 99.1/\textbf{29.5} & 85.3/40.1 & \textbf{99.9}/\textbf{29.7} \\
    \bottomrule
\end{tabular}
\label{table:model robust}
\end{table}

For model-level attacks, we consider finetuning and model compression, namely pruning, and quantization. In the finetuning experiments, we perform 20k iterations of the GAN fingerprinting model on the same dataset used for training by removing the watermark losses from the optimization, that is, setting $\lambda_1 = 1$ and $\lambda_i = 0$, $i = 2,3, 4$,  and leaving the parameters of the PN layer free to update. For pruning, we set to 0 the smallest $p\%$ parameters of the network, with $ p = 10\%$ and $20\%$. Finally, for quantization, we reduce the precision of the model parameters by rounding them to the first digit.

The results we got are shown in Table~\ref{table:model robust}. We can observe that, for all the architectures, the proposed approach is more robust against finetuning compared to \cite{DBLP:conf/iclr/YuSCDF22} and~\cite{fei2022supervised}, especially in the case of element-wise PN, where Acc is 20\% higher with respect to the state-of-the-art on the average. All approaches are robust against pruning and quantization. Pruning with $p = 20\%$ can affect the watermark yet at the price of a very large FID, corresponding to a bad quality of the generated images, that makes the model useless.

Table~\ref{table:image robust} reports the results of robustness against image-level attacks, in the case of JPEG compression (with quality factor 50), Gaussian blur (with kernel size 5), and Gaussian noise (with standard deviation 0.1). We can observe that our method achieves the best robustness in all the cases, especially for JPEG compression, in which case the gain in the Acc is about 15/20\%.

\begin{table}[htbp]
    \renewcommand{\arraystretch}{1}
    \centering
    \caption{Watermark Acc under different image-level attacks}
    \begin{tabular}{c|ccccc}
    \toprule
    Model       &    Attack    & Yu~\cite{DBLP:conf/iclr/YuSCDF22}    & Fei~\cite{fei2022supervised}   &  -\textit{cw}   &  -\textit{ew}\\ \midrule
    \multirow{4}*{BEGAN} & JPEG        & 71.42   & 77.63 &  \textbf{92.62} &  92.48\\
    & Blurring    & 76.31   & 70.35 & \textbf{81.49}  &  80.37\\
    &  Noise  & 75.93  &74.17 & 88.50  &  \textbf{89.33}\\ \midrule

    \multirow{4}*{SNGAN} & JPEG        & 72.99   & 76.31 &  92.01 &  \textbf{92.17}\\
    & Blurring    & 77.17   & 72.34 & \textbf{83.23}  &  80.45\\
    &  Noise  & 74.34  &73.34 & 88.21  &  \textbf{89.52}\\ \midrule

    \multirow{4}*{PGGAN} & JPEG        & 73.15   & 78.54 &  \textbf{94.25} &  93.67\\
    & Blurring    & 74.46   & 72.54 & 80.17  &  \textbf{83.70}\\
    &  Noise  & 73.39  &76.40 & 86.14  &  \textbf{88.28}\\
    \bottomrule
\end{tabular}
\label{table:image robust}
\end{table}

\subsection{Ablation Study}
\label{Ablation study}

We carry out several experiments to investigate the impact of the loss terms and the position of the PN layer on the effectiveness of the proposed method. In these experiments, we focus on the element-wise PN embedding, which is the best performing method in terms of both image quality and watermark robustness, according to the previous results.
\subsubsection{Impact of the loss terms}
\begin{figure}[thbp]
    \centering
    \includegraphics[width=\linewidth]{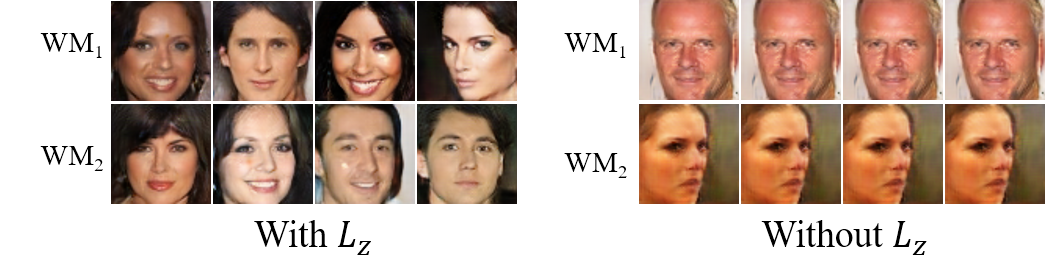}
    \caption{Images generated by the generators marked with WM$_1$ and WM$_2$ when fed with different input noises.}
     \label{fig:samples}
\end{figure}

Table~\ref{table:distangle} reports the Acc and FID obtained by training the model with and without the $L_z$ and $L_{const}$ terms and considering two generator instances corresponding to two random watermarks WM$_1$ and WM$_2$. We see that removing $L_z$ or $L_{const}$ does not affect the watermark extraction accuracy. However, removing  $L_z$ results in a very high FID, as without this loss term the content of the generated images tends to be controlled solely by the watermark, thereby reducing the diversity. Fig.~\ref{fig:samples} shows some examples of images generated by the models marked with WM$_1$ and WM$_2$, when they are trained with and without $L_z$. Although the visual quality of the generated images is good in both cases, without $L_z$, the generated images lack diversity.

\begin{table}[htbp]
    \renewcommand{\arraystretch}{1.10}
    \centering
    \caption{Performance of $G_{w1}$ and $G_{w2}$ when training with/without $L_z$ and $L_{const}$.}
\begin{tabular}{ccc}
    \toprule
    Approach          & Acc$_{\text{WM}_1}$  /  Acc$_{\text{WM}_2}$     & FID$_{\text{WM}_1}$  /  FID$_{\text{WM}_2}$  \\ \midrule
    With all losses       & 100.00 / 100.00    & 20.72 / 20.68  \\
    Without $L_z$    & 100.00 / 100.00    & 233.38 / 205.86 \\
    Without $L_{const}$  & 100.00 / 100.00   & 21.18 / 20.94  \\

    \bottomrule
\end{tabular}
\label{table:distangle}
\end{table}

\begin{figure}[thbp]
    \vspace{-0.5cm}
    \centering
    \includegraphics[width=0.9\linewidth]{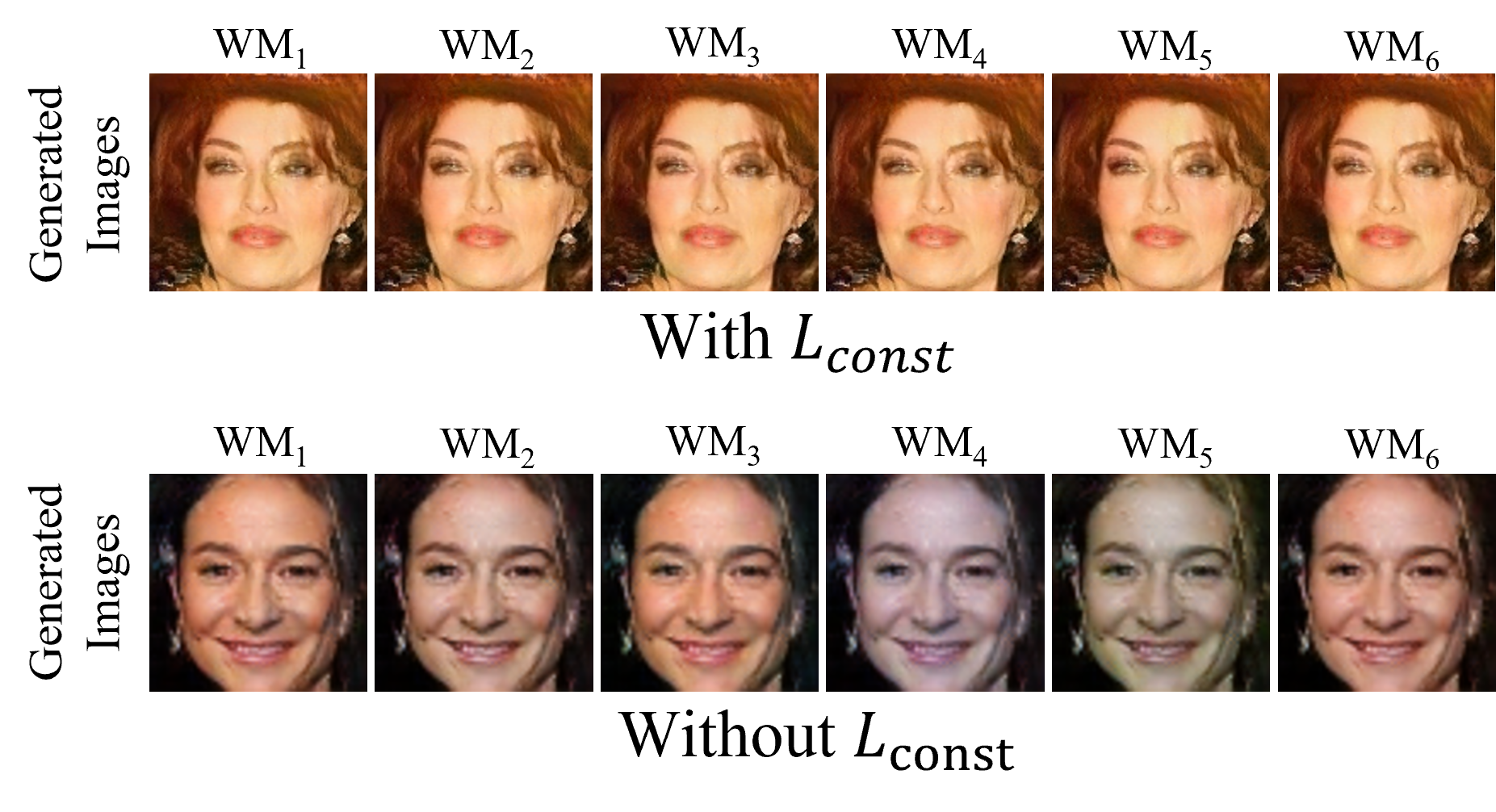}
    \caption{Images generated by generator instances with different watermarks.}
     \label{fig:samples2}
     \vspace{-0.5cm}
\end{figure}

Regarding $L_{const}$, Table~\ref{table:distangle} shows that when training is carried out without this loss term, both FID and Acc remain good. However, the purpose of this loss is to ensure that model instances with different watermarks have the same functionality, that is, that the watermarked models generate images that are visually the same when fed with the same noise. Fig.~\ref{fig:samples2} shows some images produced by 6 generator instances with different watermarks when they are fed with the same input noise $z$, in the case where training is performed with and without $L_{const}$. We can observe that without $L_{const}$, the images obtained for the same $z$ from the 6 generators are visually different. This does not happen when the training includes the $L_{const}$ loss.

\begin{table}[htbp]
    \renewcommand{\arraystretch}{1}
    \centering
    \caption{Impact of the position of the PN layer (FT = finetuning).}
\begin{tabular}{m{0.9cm}<{\centering}|m{2cm}<{\centering}|m{1.0cm}<{\centering}m{0.8cm}<{\centering}m{0.8cm}<{\centering}m{0.8cm}<{\centering}}
    \toprule
    Model        & Metric        & \makecell[c]{Input}  & \makecell[c]{Mid}     & \makecell[c]{Output\\(default)}  & \makecell[c]{All} \\ \midrule
    \multirow{3}*{BEGAN} & Acc  & \textbf{100.00}    & \textbf{100.00}  & \textbf{100.00}  & \textbf{100.00} \\
    &FID  &  21.24    & 20.89 & 20.72 & \textbf{20.24}\\
    &Acc after FT   & 68.24    & 72.68 & \textbf{75.92} & 64.46\\ \midrule

    \multirow{3}*{SNGAN} & Acc  & \textbf{99.99}   & 99.90 & \textbf{99.99} & 99.98\\
    &FID  & 24.15    & 24.76 & 24.70 & \textbf{24.01}\\
    &Acc after FT    & 72.54    & 84.32 & \textbf{88.21} & 64.24\\

    \bottomrule
\end{tabular}
\label{table:position}
\end{table}

\subsubsection{Impact of the position of the PN layer}
The previous experiments are carried out with the PN layer included as the penultimate layer (default position). In this section, we report the results of the experiments we run considering different positions for this layer. In particular, we consider the following cases: i) the PN layer is added as the second layer after the input layer (\textit{Input}); ii) the PN layer is inserted as the middle layer (\textit{Mid}); iii) multiple PN layers are included, after every convolutional block of the generator (\textit{All}). In the {\textit{All}} case, all the  ParamGen Nets have the same internal architecture and the watermark message is fed as input to all of them. For these experiments, we consider only BEGAN and SNGAN, since in the PGGAN case the dynamic growth of the generator during training complicates the implementation of the approach. In particular, we find that including the PN layer as an intermediate layer at a late stage during training (when the final output resolution is achieved) makes training unstable.

The results are reported in Table \ref{table:position}. We observe that the position of the PN layer has a few impacts on Acc and FID. However, an impact is observed on watermark robustness. In particular, the \textit{All} case is the worst from the point of view of robustness, with a reduction of 11.46\% for BEGAN and 23.97\% for SNGAN, compared to the default setting. The \textit{Mid} and penultimate layer position (\textit{Default}) are those maximizing the robustness against finetuning attacks, with the default achieving the best results.

\subsection{Discussion on collusion attacks}
\label{sec.Limitations}
\begin{figure}[thbp]
    \centering
    \includegraphics[width=0.65\linewidth]{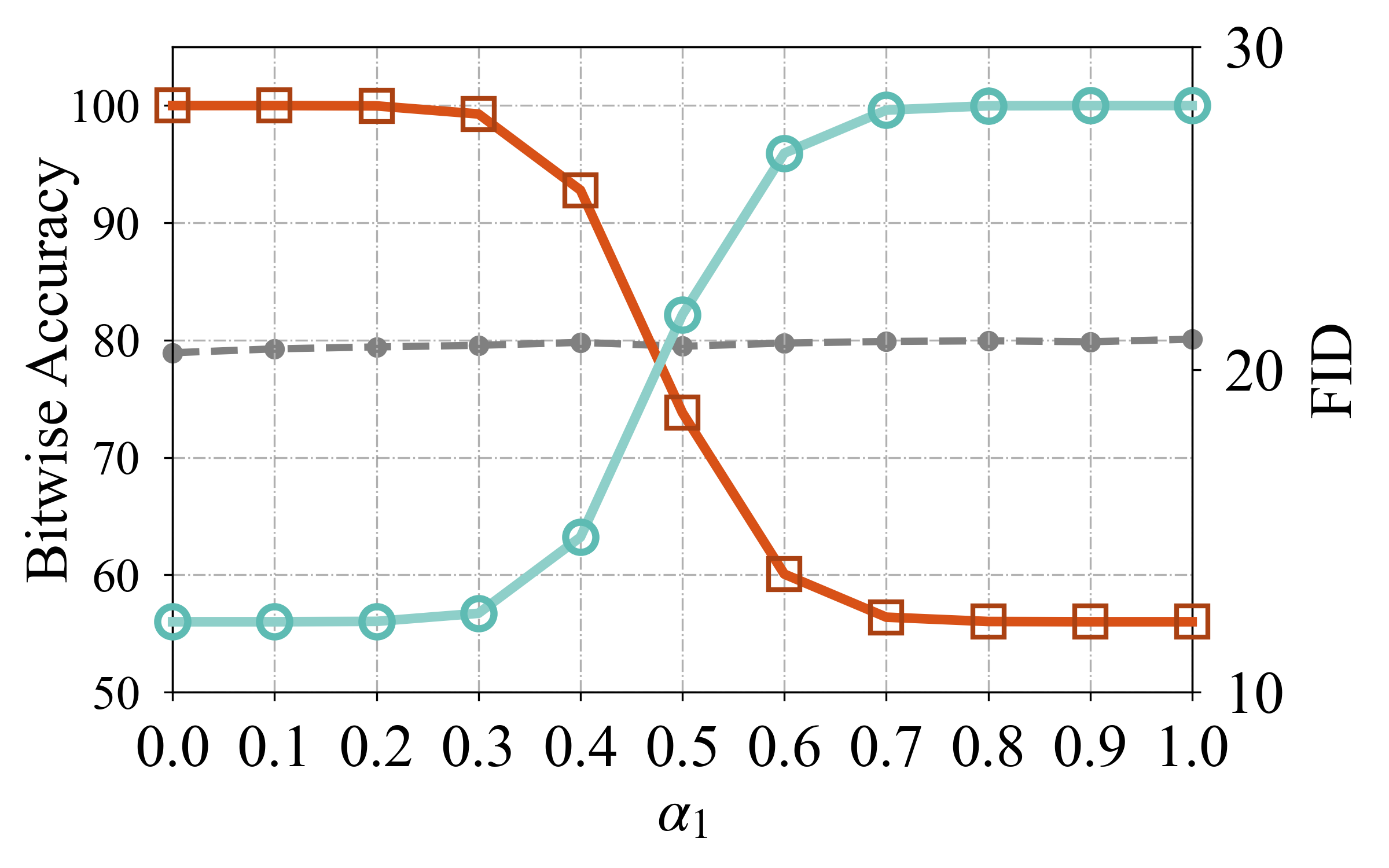}
    \caption{Watermark accuracy and FID under collusion attack. The red curve with squares (green curve with circles) represents the bit matching accuracy between the watermark extracted from $G_{attk}$ and $w_A$ ($w_B$). The gray curve with dots reports the FID of the images generated by $G_{attk}$.}
     \label{fig:coll}
\end{figure}

In this section, we pause to discuss a specific watermark removal attack, namely the collusion attack, that is particularly relevant in the GAN fingerprinting scenario. In this attack, different generator instances are combined together in order to produce a new generator that can achieve the same functionality and does not contain the watermark information of the source instances. Let $G_{w_A}$ and $G_{w_B}$ be two generator instances distributed to users $A$ and $B$ containing the user-specific fingerprints $w_A$ and $w_B$. Then, an attacker who has access to the model parameters of $G_{w_A}$ and $G_{w_B}$ can generate a new model $G_{attk} = \alpha_1 G_{w_A} + \alpha_2 G_{w_B}$, with $\alpha_1 + \alpha_2 = 1$. Since the parameters of $G_{w_A}$ and $G_{w_B}$ are the same except for the PN layers, the interpolation only affects the parameters of the PN layer. Arguably, the watermark extracted by $D_{w}$ from $G_{attk}$ would contain only part of the watermark information of $w_A$ and $w_B$. Fig.~\ref{fig:coll} shows the bit matching accuracy between the watermark extracted from $G_{attk}$ and, respectively, $w_A$ and $w_B$, for different values of $\alpha_1$. The FID, also reported in the figure, is approximately constant over $\alpha_1$. We observe that, in the case where $\alpha_1 = 0.5$, the correlation of the extracted watermark with the two watermarks remains high, and Acc is around 75\% and 83\% for watermark $w_A$ and $w_B$ respectively. Please note that watermark A and B are randomly selected once. With the increase in the number of experiments, when $\alpha_1 = 0.5$, the Acc for both watermarks will approach 50\%.
In this condition, the collusion attack can be mitigated and the Acc recovered by incorporating error correction or traitor tracing codes. Since the number of distinct model instances that can be produced is extremely large (namely, $3.4 \times 10^{38}$ with the payload of $d_w=128$ bits considered in this paper), a fraction of bits could be utilized to perform error correction to prevent collusion by malicious users.

\section{Conclusions}
\label{Conclusions}
We have proposed a robust retraining-free GAN fingerprinting system that allows robust box-free watermarking of GAN generators, making it easy for the model owner to distribute copies of the generator with the same functionality but different watermarks (user-specific fingerprints).
According to the experiments we carry out considering several architectures, the proposed method achieves very good results, always overcoming the state-of-the-art in terms of robustness against model-level and image-level attacks. Future work will focus on the extension of the proposed method to different generative architectures, e.g., diffusion models, also considering images belonging to different domains. Investigating the use of channel coding and error correction codes to increase the robustness against watermark removal and collusion attacks is also a very interesting topic for future research.

\section*{Acknowledgment}
This work is supported in part by the National Key R\&D Program of China under Grant number 2022YFB3103100, in part by the National Natural Science Foundation of China under grant numbers 62122032, 62172233, 62102189. This work has been partially supported by the China Scholarship Council (CSC), file No.202109040029.


{
    \small
    \bibliographystyle{IEEEtran}
    \bibliography{IEEEexample}

\begin{thebibliography}{10}
\providecommand{\url}[1]{#1}
\csname url@samestyle\endcsname
\providecommand{\newblock}{\relax}
\providecommand{\bibinfo}[2]{#2}
\providecommand{\BIBentrySTDinterwordspacing}{\spaceskip=0pt\relax}
\providecommand{\BIBentryALTinterwordstretchfactor}{4}
\providecommand{\BIBentryALTinterwordspacing}{\spaceskip=\fontdimen2\font plus
\BIBentryALTinterwordstretchfactor\fontdimen3\font minus
  \fontdimen4\font\relax}
\providecommand{\BIBforeignlanguage}[2]{{%
\expandafter\ifx\csname l@#1\endcsname\relax
\typeout{** WARNING: IEEEtran.bst: No hyphenation pattern has been}%
\typeout{** loaded for the language `#1'. Using the pattern for}%
\typeout{** the default language instead.}%
\else
\language=\csname l@#1\endcsname
\fi
#2}}
\providecommand{\BIBdecl}{\relax}
\BIBdecl

\bibitem{barni2021dnn}
M.~Barni, F.~P{\'e}rez-Gonz{\'a}lez, and B.~Tondi, ``Dnn watermarking: four
  challenges and a funeral,'' in \emph{Proceedings of the 2021 ACM Workshop on
  Information Hiding and Multimedia Security}, 2021, pp. 189--196.

\bibitem{ong2021protecting}
D.~S. Ong, C.~S. Chan, K.~W. Ng, L.~Fan, and Q.~Yang, ``Protecting intellectual
  property of generative adversarial networks from ambiguity attacks,'' in
  \emph{Proceedings of the IEEE/CVF Conference on Computer Vision and Pattern
  Recognition}, 2021, pp. 3630--3639.

\bibitem{DBLP:conf/iclr/YuSCDF22}
N.~Yu, V.~Skripniuk, D.~Chen, L.~S. Davis, and M.~Fritz, ``Responsible
  disclosure of generative models using scalable fingerprinting,'' in \emph{The
  Tenth International Conference on Learning Representations, {ICLR} 2022,
  Virtual Event, April 25-29, 2022}, 2022.

\bibitem{zhang2020passport}
J.~Zhang, D.~Chen, J.~Liao, W.~Zhang, G.~Hua, and N.~Yu, ``Passport-aware
  normalization for deep model protection,'' \emph{Advances in Neural
  Information Processing Systems}, vol.~33, pp. 22\,619--22\,628, 2020.

\bibitem{li2021survey}
Y.~Li, H.~Wang, and M.~Barni, ``A survey of deep neural network watermarking
  techniques,'' \emph{Neurocomputing}, vol. 461, pp. 171--193, 2021.

\bibitem{wu2020watermarking}
H.~Wu, G.~Liu, Y.~Yao, and X.~Zhang, ``Watermarking neural networks with
  watermarked images,'' \emph{IEEE Transactions on Circuits and Systems for
  Video Technology}, vol.~31, no.~7, pp. 2591--2601, 2020.

\bibitem{yu2021artificial}
N.~Yu, V.~Skripniuk, S.~Abdelnabi, and M.~Fritz, ``Artificial fingerprinting
  for generative models: Rooting deepfake attribution in training data,'' in
  \emph{Proceedings of the IEEE/CVF International Conference on Computer
  Vision}, 2021, pp. 14\,448--14\,457.

\bibitem{fei2022supervised}
J.~Fei, Z.~Xia, B.~Tondi, and M.~Barni, ``Supervised gan watermarking for
  intellectual property protection,'' in \emph{2022 IEEE International Workshop
  on Information Forensics and Security (WIFS)}.\hskip 1em plus 0.5em minus
  0.4em\relax IEEE, 2022, pp. 1--6.

\bibitem{qiao2023novel}
T.~Qiao, Y.~Ma, N.~Zheng, H.~Wu, Y.~Chen, M.~Xu, and X.~Luo, ``A novel model
  watermarking for protecting generative adversarial network,'' \emph{Computers
  \& Security}, p. 103102, 2023.

\bibitem{quan2020watermarking}
Y.~Quan, H.~Teng, Y.~Chen, and H.~Ji, ``Watermarking deep neural networks in
  image processing,'' \emph{IEEE transactions on neural networks and learning
  systems}, vol.~32, no.~5, pp. 1852--1865, 2020.

\bibitem{tancik2020stegastamp}
M.~Tancik, B.~Mildenhall, and R.~Ng, ``Stegastamp: Invisible hyperlinks in
  physical photographs,'' in \emph{Proceedings of the IEEE/CVF Conference on
  Computer Vision and Pattern Recognition}, 2020, pp. 2117--2126.

\bibitem{srivastava2017veegan}
A.~Srivastava, L.~Valkov, C.~Russell, M.~U. Gutmann, and C.~Sutton, ``Veegan:
  Reducing mode collapse in gans using implicit variational learning,''
  \emph{Advances in neural information processing systems}, vol.~30, 2017.

\bibitem{berthelot2017began2}
D.~Berthelot, T.~Schumm, and L.~Metz, ``Began: Boundary equilibrium generative
  adversarial networks,'' \emph{arXiv preprint arXiv:1703.10717}, 2017.

\bibitem{miyatospectral}
T.~Miyato, T.~Kataoka, M.~Koyama, and Y.~Yoshida, ``Spectral normalization for
  generative adversarial networks,'' in \emph{International Conference on
  Learning Representations}.

\bibitem{karrasprogressive}
T.~Karras, T.~Aila, S.~Laine, and J.~Lehtinen, ``Progressive growing of gans
  for improved quality, stability, and variation,'' in \emph{International
  Conference on Learning Representations}.

\bibitem{liu2015deep}
Z.~Liu, P.~Luo, X.~Wang, and X.~Tang, ``Deep learning face attributes in the
  wild,'' in \emph{Proceedings of the IEEE international conference on computer
  vision}, 2015, pp. 3730--3738.

\bibitem{heusel2017gans}
M.~Heusel, H.~Ramsauer, T.~Unterthiner, B.~Nessler, and S.~Hochreiter, ``Gans
  trained by a two time-scale update rule converge to a local nash
  equilibrium,'' \emph{Advances in neural information processing systems},
  vol.~30, 2017.

\end{thebibliography}
}

\end{document}